\documentclass[11pt,a4paper]{article}
\usepackage[hyperref]{naacl2021}
\usepackage{times}
\usepackage{latexsym}

\usepackage[T5,T2A,T1]{fontenc}
\usepackage[utf8]{inputenc}

\usepackage{url}
\usepackage{nicefrac}
\usepackage{calc}
\usepackage{xcolor}
\usepackage{graphicx}
\usepackage{multirow}

\usepackage{amsthm}
\usepackage{amssymb}
\usepackage{amsmath}
\usepackage{amsfonts}
\usepackage[toc]{appendix}
\usepackage{color, colortbl}
\usepackage{adjustbox}
\usepackage{tabularx}
\usepackage{booktabs}
\usepackage{subcaption}
\usepackage{makecell}
\usepackage{arydshln}
\usepackage{tikz}
\usepackage{enumitem}
\usetikzlibrary{positioning}

\definecolor{Gray}{gray}{0.92}
\newcolumntype{Y}{>{\centering\arraybackslash}X}

\usepackage{microtype}

\usepackage{todonotes}
\makeatletter
\newcommand*\iftodonotes{\if@todonotes@disabled\expandafter\@secondoftwo\else\expandafter\@firstoftwo\fi}
\makeatother

\title{\textit{ConVEx}: Data-Efficient and Few-Shot Slot Labeling}

\author{Matthew Henderson$^{\mathbf{1}}$ and Ivan Vuli\'{c}$^{\mathbf{1,2}}$\\
$^{\mathbf{1}}$  PolyAI Ltd, London, UK\\
$^{\mathbf{2}}$ Language Technology Lab, University of Cambridge, UK\\
{\texttt{ivan@polyai.com}}}

\date{}

\begin{document}
\maketitle
\begin{abstract}
We propose \textbf{ConVEx} (\textbf{Con}versational \textbf{V}alue \textbf{Ex}tractor), an efficient pretraining and fine-tuning neural approach for slot-labeling dialog tasks. Instead of relying on more general pretraining objectives from prior work (e.g., language modeling, response selection), ConVEx's pretraining objective, a novel \textit{pairwise cloze} task using Reddit data, is well aligned with its intended usage on sequence labeling tasks. This enables learning domain-specific slot labelers by simply fine-tuning decoding layers of the pretrained general-purpose sequence labeling model, while the majority of the pretrained model's parameters are kept frozen. We report state-of-the-art performance of ConVEx across a range of diverse domains and data sets for dialog slot-labeling, with the largest gains in the most challenging, few-shot setups. We believe that ConVEx's reduced pretraining times (i.e., only 18 hours on 12 GPUs) and cost, along with its efficient fine-tuning and strong performance, promise wider portability and scalability for data-efficient sequence-labeling tasks in general.
\end{abstract}

\section{Introduction}
\label{s:intro}
% P1: what is slot labeling? why is it important for dialog?
Slot labeling or slot filling is a critical natural language understanding (NLU) component of any task-oriented dialog system \cite[\textit{inter alia}]{Young:02is, young:10b, Tur:2011}. Its goal is to fill the correct \textit{values} associated with predefined \textit{slots}: e.g., a dialog system for restaurant bookings is expected to fill slots such as \textit{date}, \textit{time}, and \textit{the number of guests} with the values extracted from a user utterance (e.g., \textit{next Thursday}, \textit{7pm}, \textit{4 people}).

%% P2: why do we need data-efficient learning?
Setting up task-oriented dialog systems, as well as slot labeling methods in particular, to support new tasks and domains is highly challenging due to inherent scarcity of expensive expert-annotated data for a plethora of intended use scenarios \cite{Williams:14,Henderson:14a,Budzianowski:2018emnlp,Zhao:2019naacl}. One plausible and promising solution is the creation of \textit{data-efficient} models that learn from only a handful annotated examples in \textit{few-shot scenarios}. This approach has been shown promising for learning intent detectors \cite{Casanueva:2020ws,krone2020,Bunk:2020arxiv} as well as for slot-filling methods \cite{hou2020few,CoopeFarghly2020}. %with limited annotated data.

%% P3: The current SOTA and the main gaps and challenges
The dominant paradigm followed by the existing models of few-shot slot labeling is \textit{transfer learning} \cite{Ruder:2019transfer}: \textbf{1)} they rely on representations from models \textit{pretrained} on large data collections in a self-supervised manner on some general NLP tasks such as (masked) language modeling \cite{Devlin:2018arxiv,Conneau:2020acl,Brown:2020gpt3} or response selection \cite{Henderson:2019acl,Henderson:2019convert,Cer:2018arxiv}; and then \textbf{2)} add additional task-specific layers for modeling the input sequences. However, we detect several gaps with the existing setup, and set to address them in this work. First, recent work in NLP has validated that a stronger alignment between a pretraining task and an end task can yield performance gains for tasks such as extractive question answering \cite{glass2020span} and paraphrase and translation \cite{Lewis:2020marge}. We ask whether it is possible to design a pretraining task which is more suitable for slot labeling in conversational applications. Second, is it possible to bypass learning sequence-level layers from scratch, and simply \textit{fine-tune} them after pretraining instead?  Third, is it possible to build a generally applicable model which fine-tunes pretrained ``general'' sequence-level layers instead of requiring specialized slot labeling algorithms from prior work \cite{krone2020,hou2020few}?

%% P4: UVE and our main contributions
Inspired by these challenges, we propose \textbf{ConVEx} (\textbf{Con}versational \textbf{V}alue \textbf{Ex}tractor), a novel Transformer-based neural model which can be pretrained on large quantities of natural language data (e.g., Reddit) and then directly fine-tuned to a variety of slot-labeling tasks. Similar to prior work \cite{rastogi2019towards,CoopeFarghly2020}, ConVEx casts slot labeling as a span-based extraction task. For ConVEx, we introduce a new pretraining objective, termed \textit{pairwise cloze}. This objective aligns well with the target downstream task: slot labeling for dialog, and emulates slot labeling relying on unlabeled sentence pairs from natural language data which share a keyphrase (i.e., a ``value'' for a specific ``slot''). Instead of learning them from scratch as in prior work \cite{CoopeFarghly2020}, ConVEx's pretrained Conditional Random Fields (CRF) layers for sequence modeling are fine-tuned using a small number of labeled in-domain examples. 

We evaluate ConVEx on a range of diverse dialog slot labeling data sets spanning different domains: \textsc{dstc8} data sets \cite{rastogi2019towards}, \textsc{restaurants-8k} \cite{CoopeFarghly2020}, and SNIPS \cite{coucke2018snips}. ConVEx yields state-of-the-art performance across all evaluation data sets, but its true usefulness and robustness come to the fore in the few-shot scenarios. For instance, it increases average $F_1$ scores on \textsc{restaurants-8k} over the previous state-of-the-art model \cite{CoopeFarghly2020} from 40.5 to 71.7 with only 64 labeled examples. Similar findings are observed with \textsc{dstc8}, and we also report state-of-the-art performance in the 5-shot slot labeling task on SNIPS.

In summary, our results validate the benefits of \textit{task-aligned pretraining from raw natural language data}, with particular gains for data-efficient slot labeling given a limited number of annotated examples, which is a scenario typically met in production. They also clearly demonstrate that competitive performance can be achieved via quick fine-tuning, without heavily engineered specialized methods from prior work \cite{hou2020few}. Further, we validate that learning sequence-level layers from scratch is inferior to fine-tuning from pretrained layers. From a broader perspective, we hope that this research will inspire further work on task-aligned pretraining objectives for other NLP tasks beyond slot labeling. From a more focused perspective, we hope that it will guide new approaches to data-efficient slot labeling for dialog.

% why is few-shot interesting?
%\cite{hou2020few, krone2020, CoopeFarghly2020} 
% trend of pre-trained models being applied to small-data tasks. specifically trend of aligning the pre-trained model with the eventual task

% existing methods for data-efficient slot labelling take representations learned on other NLP tasks, and add layers for modelling the sequence.
% typically because it is hard to find high quantities of slot-labeling data. some sources include wikipedia data sets. 
% some motivation: it's easier to fine-tune layers than to learn them from scratch. also if a pretraining task is more similar to the downstream task the better.
% we try to find a way to use any natural language data and turn it into a slot-labeling type data set
% sequence-level layers are more complex than e.g. hidden layers of intent classifier- so we expect to benefit if those layers can be pretrained rather than learned from scratch (as in span-convert, span-bert)

% contributions of this paper:
% - new pretraining task
% - new model, uve
% - data-efficient fine-tuning strategy. no new layers to learn, just fine-tuning pre-trained parameters. competitive on few-shot without requiring specialised learning algorithms beyond fine-tuning

\section{Methodology}
\label{s:methodology}
\begin{table*}[t!]
\def\arraystretch{1.3}
{\small
\begin{center}
\resizebox{1\textwidth}{!}{% <------ Don't forget this %
\begin{tabular}{p{10.5cm} p{10.5cm}}
\toprule
\textbf{\textit{Template} Sentence} & \textbf{\textit{Input} Sentence} \\
\cmidrule(lr){1-1} \cmidrule(lr){2-2}
I get frustrated everytime I browse /r/all. I stick to my  \textit{BLANK}  most of the time. &	/r/misleadingpuddles  Saw it on the \textbf{frontpage}, plenty of content if you like the premise. \\
Why Puerto Rico? It’s Memphis at Dallas, which is in Texas where  \textit{BLANK}  hit &	\textbf{Hurricane Harvey}. Just a weird coincidence. \\
\textit{BLANK}  is my 3rd favorite animated Movie	& \textbf{Toy Story 3} ended perfectly, but Disney just wants to keep milking it. \\
It really sucks, as the V30 only has  \textit{BLANK} . Maybe the Oreo update will add this. &	Thanks for the input, but \textbf{64GB} is plenty for me :) \\
I took  \textit{BLANK}, cut it to about 2 feet long and duct taped Vive controllers on each end. Works perfect &	Yeah, I just duct taped mine to \textbf{a broom stick}. You can only play no arrows mode but it’s really fun. \\
I had  \textit{BLANK}  and won the last game and ended up with 23/20 and still didn’t get it. &	I know how you feel my friend and I got \textbf{19/20} on the tournament today \\
\bottomrule
\end{tabular}
}%
\end{center}
}
\vspace{-0.5mm}
\caption{Sample data from Reddit converted to sentence pairs for the ConVEx pretraining via the pairwise cloze task. Target spans in the input sentence are denoted with bold, and are \textit{``BLANKed''} in the template sentence.}
\label{tab:redditSample}
\vspace{-0.5mm}
\end{table*}

Before we delve deeper into the description of ConVEx in \S\ref{ss:uve}, in \S\ref{ss:task} we first describe a novel sentence-pair value extraction pretraining task used by ConVEx, called \emph{pairwise cloze}, and then in \S\ref{ss:data} a procedure that converts ``raw'' unlabeled natural language data into training examples. %for the ConVEx pairwise cloze task. 

%\matt{Ivan, do you think we could adapt something like the first paragraph of the `Pairwise Cloze Pretraining' section in the blog post? - i.e. say we want to come up with something that is better aligned, and allows pre-training all the layers.}

\subsection{Pretraining Task: Pairwise Cloze}
\label{ss:task}
\noindent \textbf{Why Pairwise Cloze?} Top performing natural language understanding models typically make use of neural nets pretrained on large scale data sets with unsupervised objectives such as language modeling \cite{Devlin:2018arxiv,Liu:2019roberta} or response selection \cite{Henderson:2019convert,Humeau:2019arxiv}. For sequential tasks such as slot labeling, this involves adding new layers and training them from scratch, as the pretraining procedure does not involve any sequential decoding; therefore, current unsupervised pretraining objectives are suboptimal for sequence-labeling tasks. With ConVEx, we introduce a new pretraining task with the following properties: \textbf{1)} it is more closely related to the target slot-labeling task, and \textbf{2)} it facilitates training all the necessary layers for slot-labeling, so these can be fine-tuned rather than learned from scratch.

%The pretraining of ConVEx relies on a novel \textit{pairwise cloze} task. 

\vspace{1.8mm}
\noindent \textbf{What is Pairwise Cloze?} In a nutshell, given a pair of sentences that have a \textit{keyphrase} in common, the task treats one sentence as a \textit{template sentence} and the other as its corresponding \textit{input sentence}. For the template sentence, the keyphrase is masked out and replaced with a special \textit{BLANK} token. The model must then read the tokens of both sentences, and predict which tokens in the input sentence constitute the masked phrase.  Some examples of such pairs extracted from Reddit are provided in Table~\ref{tab:redditSample}. The main idea is to teach the model an implicit space of \textit{slots} and \textit{values}, where during self-supervised pretraining, slots are represented as the contexts in which a value might occur. The model than gets fine-tuned later to fit domain-specific slot labeling data.\footnote{The pairwise cloze task has been inspired by the recent span selection objective applied to extractive QA by \newcite{glass2020span}: they create examples emulating extractive QA pairs with long passages and short question sentences. Another similar approach to extractive QA has been proposed by \newcite{Ram:2021span}. In contrast, our work seeks to emulate slot labeling in a dialog system by creating examples from short conversational utterances.}

\subsection{Pairwise Cloze Data Preparation}
\label{ss:data}
%\vspace{1.4mm}
\noindent \textbf{Input Data.} 
We assume working with the English language throughout the paper. Reddit has been shown to provide natural conversational English data for learning semantic representations that work well in downstream tasks related to dialog and conversation \cite{AlRfou:2016arxiv,Cer:2018arxiv,Henderson:2019acl,Henderson:2019arxiv,Henderson:2019convert,Casanueva:2020ws,CoopeFarghly2020}. Therefore, following recent work, we start with the 3.7B comments in the large Reddit corpus from 2015-2018 (inclusive) \cite{Henderson:2019arxiv}, filtering it to comments between 9 and 127 characters in length. This yields a total of almost 2B filtered comments.

\vspace{1.8mm}
\noindent \textbf{Keyphrase Identification.} Training sentence pairs are extracted from unlabeled text based on their shared keyphrases. Therefore, we must first identify plausible candidate keyphrases. To this end, the filtered Reddit sentences are tokenized with a simple word tokenizer, and word frequencies are counted. The score of a candidate keyphrase $kp = \left(w_1,\:w_2,\:\ldots,w_n\right)$ is computed as a function of the individual word counts:
%
% \begin{equation}
% score(kp) = \Big(\prod_{i=1}^{n} \mathrm{count}(w_i) \Big)^{-\nicefrac{1}{n^\alpha}}.
% \end{equation}
\begin{equation}
 score(kp) = \nicefrac{1}{n^\alpha} \sum_{i=1}^{n} \log  \frac{|D|}{\mathrm{count}(w_i)}.
\end{equation}
%
%% \footnote{In practice the log score is easier to compute.}
%
\noindent where $|D|$ is the number of sentences used to calculate the word frequencies. This simple scoring function selects phrases that have informative low-frequency words. The factor $\alpha$ controls the length of the identified keyphrases: e.g., setting it to $\alpha=0.8$, which is default in our experiments later, encourages selecting longer phrases. Given a sentence, the keyphrases are selected as those unigrams, bigrams and trigrams whose score exceeds a predefined threshold.

The keyphrase identification procedure is run for all sentences from the filtered Reddit sentences. At most two keyphrases are extracted per sentence, and keyphrases spanning more than 50\% of the sentence text are ignored. Keyphrases that occur more than once in the sentence are also ignored.

\vspace{1.8mm}
\noindent \textbf{Sentence-Pair Data Extraction.}
In the next step, sentences from the same subreddit are paired by keyphrase to create paired data, 1.2 billion examples in total,\footnote{We also expand keyphrases inside paired sentences if there is additional text on either side of the keyphrase that is the same in both sentences. For instance, the original keyphrase \textit{``Star Wars''} will be expanded to the keyphrase \textit{``Star Wars movie''} within this pair: \textit{``I really enjoyed the latest Star Wars movie.''} -- \textit{``We could not stand any Star Wars movie.''}} where one sentence acts as the input sentence and another as the template sentence (see Table~\ref{tab:redditSample} again). Table~\ref{tab:redditData} summarizes statistics from the entire pretraining data preparation procedure.

\begin{table}[t!]
\def\arraystretch{1.09}
%\resizebox{0.47\textwidth}{!}
{%
\small
\begin{center}
\begin{tabular}{l r}
\toprule
Total Reddit comments & 3,680,746,776 \\
Comments filtered by length & 1,993,294,538 \\
Extracted keyphrases & 3,296,519,827 \\
Training set size & 1,172,174,919 \\
Test set size & 61,696,649 \\
Mean number of words per keyphrase & 1.3 \\
\bottomrule
\end{tabular}
\end{center}
}
\vspace{-1.5mm}
\caption{Statistics of the pairwise cloze training data.}
\label{tab:redditData}
\vspace{-1.5mm}
\end{table}

\begin{figure*}[!t]
    \centering
    \begin{subfigure}[!t]{0.548\linewidth}
        \centering
        \includegraphics[width=1.0\linewidth]{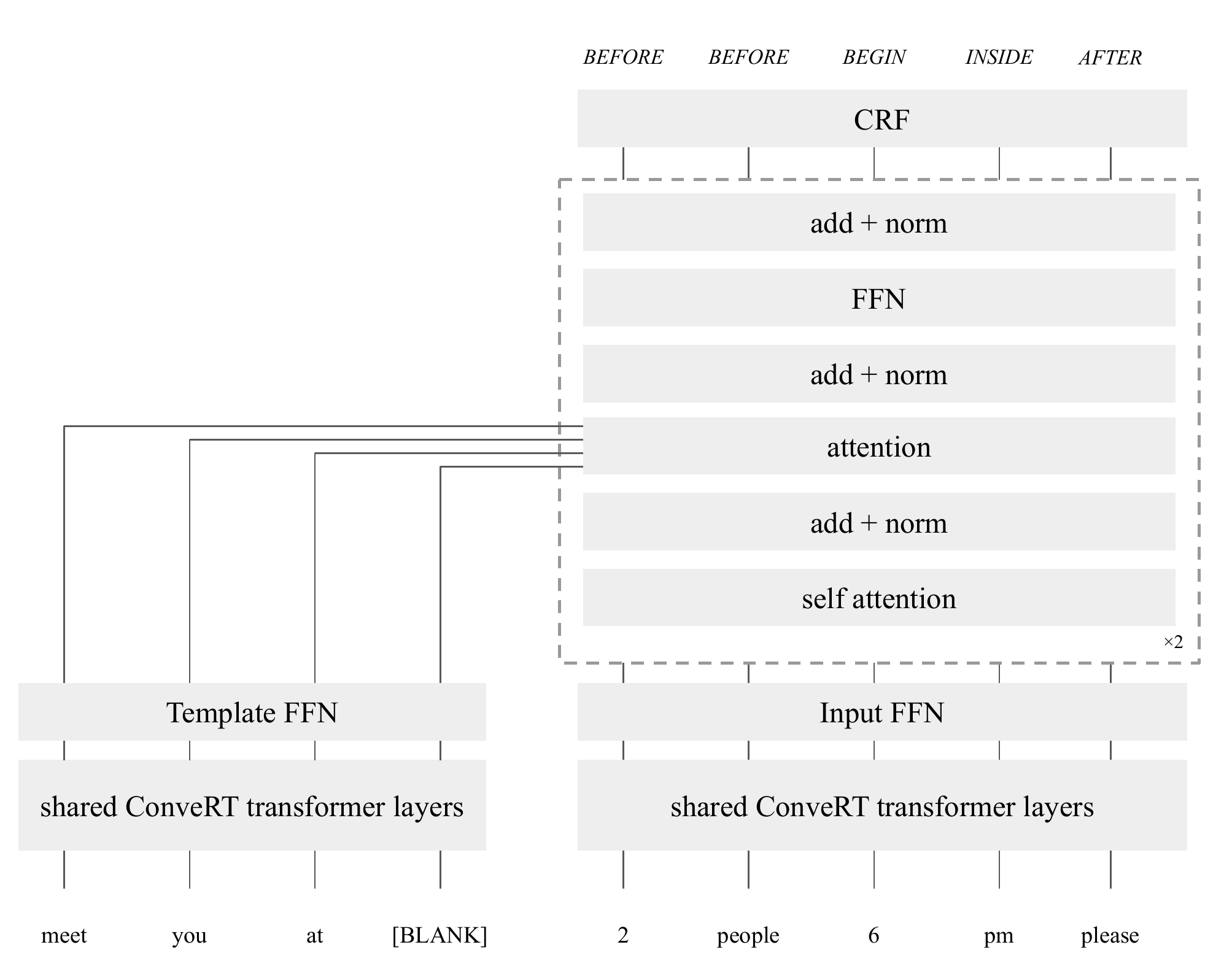}
        %\vspace{-0.7em}
        \caption{ConVEx: Pretraining}
        \label{fig:uvePretraining}
    \end{subfigure}
    \begin{subfigure}[!t]{0.428\textwidth}
        \centering
        \includegraphics[width=1.0\linewidth]{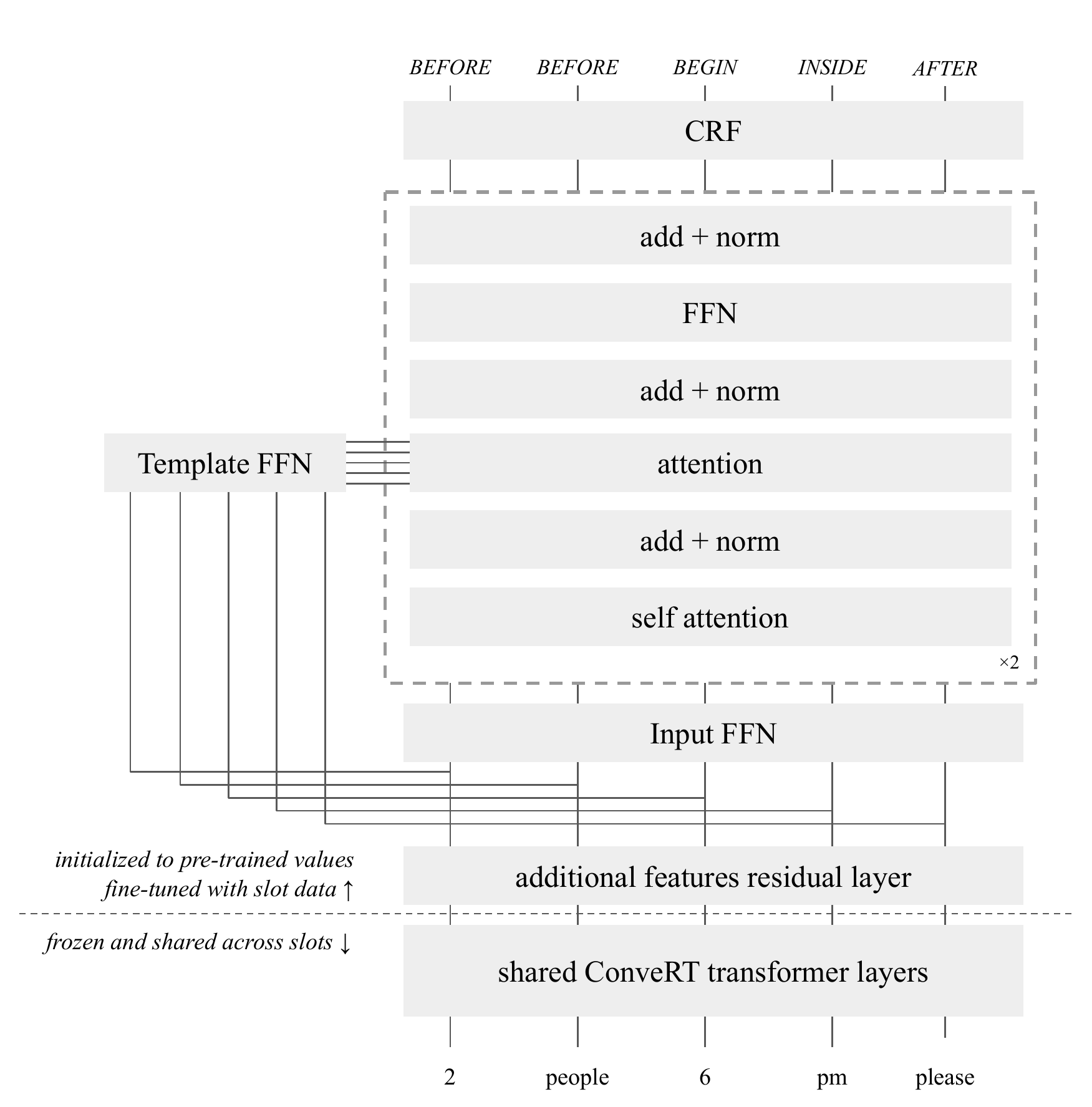}
        %\vspace{-0.7em}
        \caption{ConVEx: Fine-tuning}
        \label{fig:uveFT}
    \end{subfigure}
    \vspace{-0.5mm}
    \caption{An overview of the ConVEx model structure at: \textbf{(a)} pretraining, and \textbf{(b)} fine-tuning. The full description of each component of the model at both stages is provided in \S\ref{ss:uve}.}
    \vspace{-0.5mm}
\label{fig:uveAll}
\end{figure*}

\subsection{The ConVEx Framework}
\label{ss:uve}

We now present \textit{ConVEx}, a pretraining and fine-tuning framework that can be applied to a wide spectrum of slot-labeling tasks. ConVEx is pretrained on the pairwise cloze task (\S\ref{ss:task}), relying on sentence-pair data extracted from Reddit (\S\ref{ss:data}). Similar to prior work \cite{CoopeFarghly2020}, we frame slot labeling as a span extraction task: spans are represented using a sequence of \textit{tags}. These tags indicate which members of the sequence are in the span. We use the same tag representation as \newcite{CoopeFarghly2020}, which is similar to the standard IOB format: the span is annotated with a sequence of \textit{BEFORE}, \textit{BEGIN}, \textit{INSIDE} and \textit{AFTER} tags. The ConVEx pretraining and fine-tuning architectures are illustrated in Figures~\ref{fig:uvePretraining} and~\ref{fig:uveFT} respectively, and we describe them in what follows. 
%% IV, said before: "The overall structure of the UVE model at pretraining is given in Figure~\ref{fig:uvePretraining}."

\vspace{1.8mm}
\noindent \textbf{ConVEx: Pretraining.} The ConVEx model encodes the template and input sentences using exactly the same Transformer layer architecture \cite{Vaswani:2017nips} as the lightweight and highly optimized ConveRT sentence encoder \cite{Henderson:2019convert}: we refer the reader to the original work for all architectural and technical details. This model structure is very compact and resource-efficient (i.e., it is 59MB in size and can be trained in 18 hours on 12 GPUs) while achieving state-of-the-art performance on a range of conversational tasks \cite{Casanueva:2020ws,CoopeFarghly2020,Bunk:2020arxiv}. The weights in the ConveRT Transformer layers are shared for both sentences.\footnote{The ConVEx pretraining also closely follows ConveRT's tokenization process: the final subword vocabulary contains 31,476 subword tokens plus 1,000 buckets reserved for out-of-vocabulary tokens. Input text is split into subwords following a simple left-to-right greedy prefix matching \cite{vaswani2018,Henderson:2019convert}, and we tokenize both input sentences and template sentences the same way.}

The 512-dimensional output representations from the ConveRT layers are projected down to 128-dimensional representations using two separate feed-forward networks (FFNs), one for the template and one for the input sentence. The projected contextual subword representations of the input sentence are then enriched using two blocks of self-attention, attention over the projected template sentence representations, and FFN layers. This provides features for every token in the input sentence that take into account the context of both the input sentence and the template sentence. A final linear layer computes Conditional Random Field (CRF) parameters for tagging the value span using the 4 \textit{BEFORE}, \textit{BEGIN}, \textit{INSIDE}, and \textit{AFTER} labels. 

% universal value extractor model structure
%%\begin{figure}[tbp]
%%\textbf{\small(a) UVE Pre-training procedure}
%%    
%%\begin{center}
%%    \includegraphics[width=80mm]{img/uvePretrain.pdf}
%%\end{center}    
%%    
%%\textbf{\small(b) UVE Fine-tuning procedure}
%%\begin{center}
%%    \includegraphics[width=75mm]{img/uveFinetune.pdf}
%%\end{center}   
%%    \caption{ \label{fig:uveStructure}
%%    }
%%\end{figure}

More formally, for each step $t$, corresponding to a subword token in the input sentence, the network outputs a $4 \times 4$ matrix of transition scores $\mathbf{W}_t$ and a $4$-dimensional vector of unary potentials $\mathbf{u}_t$. Under the CRF model, the probability of a predicted tag sequence \(\mathbf{y}\) is then computed as:
%
%\vspace{-2.5mm}
%\vspace{-1em}
%{\normalsize
\begin{align*}
    p(\mathbf{y}| \mathbf{v}) \propto  \prod_{t=1}^{T-1}{
        \exp\left(
            \mathbf{W}_t |{y_{t+1},y_{t}}
        \right)}
        \prod_{t=1}^{T}{
        \exp \left( 
            \mathbf{u}_t |{y_{t}}
        \right)
        }
\end{align*}
%}%
%\vspace{-4.5mm}
%
\noindent The loss is the negative log-likelihood, which is equal to the negative sum of the transition scores and unary potentials that correspond to the true tag labels, up to a normalization term. The top scoring tag sequences are computed efficiently using the Viterbi algorithm \cite{Sutton:2012crf}.

%%\noindent \textbf{Pre-training.}
%% IV (moved to the experimental setup) The parameters of the UVE are randomly initialized (including the ConveRT layers), and the model is pre-trained on the sentence-pair value extraction Reddit data. Batches of 256 examples are constructed, 64 of which are randomly paired sentences where no value should be extracted, the remaining being paired examples from the data. This teaches the model that sometimes no value should be predicted, as is the case with slot labeling. Table~\ref{tab:hparams} gives a summary of the pre-training hyper-parameters.

In addition to the CRF loss, an \textit{auxiliary dot-product loss} can be added. This loss encourages the model to pair template sentences with the corresponding (semantically similar) input sentences. Let $\mathbf{f}^T_i$ be the $d$-dimensional encoding of the beginning-of-sentence (BOS) token for the $i^\mathrm{th}$ template sentence, and $\mathbf{f}^I_i$ be the encoding of the BOS token for the $i^\mathrm{th}$ (corrresponding) input sentence. As the encodings are contextual, the BOS representations can encapsulate the entire sequence. The auxiliary dot-product loss is then computed as:
%
%\vspace{-1.5mm}
%{\small
\begin{equation}
    -\sum_{i=1}^{N} C \left<\mathbf{f}^T_i,\:\mathbf{f}^I_i\right>
    + \sum_{i=1}^{N} \log \sum_{j=1}^{N} e^{C \left<\mathbf{f}^T_i,\:\mathbf{f}^I_j\right>}
    \label{eq:aux}
\end{equation}
%}%
%
\noindent where $\left<\cdot,\:\cdot\right>$ is cosine similarity and $C$ is an annealing factor that linearly increases from 0 to $\sqrt{d}$ over the first 10K training batches as in previous work \cite{Henderson:2019convert}. The auxiliary loss is inspired by the dot-product loss typically used in retrieval tasks such as response selection \cite{Henderson:2017arxiv}. Note that this loss does not necessitate any additional model parameters, and does not significantly increase the computational complexity of the pretraining procedure. Later in \S\ref{s:results} we evaluate the efficacy of pretraining with and without the auxiliary loss.

%% IV: (moved to 'Experimental Setup')
%%The UVE is trained for 18 hours on 12 Tesla K80 GPUs; this is typically sufficient to reach convergence. The total pretraining cost is roughly \$85 on Google Cloud Platform. This pretraining regime is orders of magnitude cheaper and more efficient than the prevalent pretrained NLP models such as BERT, GPT-2, XLNet, and RoBERTa \cite{Strubell:2019acl}.

\vspace{1.8mm}
\noindent \textbf{ConVEx: Fine-tuning.} The majority of the computation and parameters of ConVEx are in the shared ConveRT Transformer encoder layers: they comprise 30M parameters, while the decoder layers comprise only 800K parameters. At ConVEx fine-tuning, the shared ConveRT transformer layers are frozen: these expensive operations are shared across slots, while the fine-tuned slot-specific models are small in memory and fast to run.

To apply the ConVEx model to slot-labeling for a specific slot, the user utterance is treated both as the input sentence and the template sentence (note that at fine-tuning and inference the user input does not contain any \textit{BLANK} token) -- see Figure~\ref{fig:uveFT}. This effectively makes the attention layers in the decoder act like additional self-attention layers. For some domains, additional context features such as the binary \textit{is\_requested} feature need to be incorporated \cite{CoopeFarghly2020}: this is modeled through a residual layer that computes a term to add to the ConveRT output encoding, given the encoding itself and the additional features -- see Figure~\ref{fig:uveFT}.

We again note that, except for the residual layer, no new layers are added between pretraining and fine-tuning; this implies that the model bypasses learning from scratch any potential complicated dynamics related to the application task, and is directly applicable to various slot-labeling scenarios.

% fine-tuning hparams

%% \ivulic{Is there a difference between fine-tuning and inference? At fine-tuning you'd still BLANK the value for the slot in the template, right? At inference, there's no such thing? Do we need to make this more apparent? \matt{fine-tuning is the same as inference, there is no template sentence in fine-tuning, there is just the user text and corresponding slot labels.}}

\section{Experimental Setup}
\label{s:exp}
\begin{table}[t!]
\def\arraystretch{1.1}
%\resizebox{0.49\textwidth}{!}{%
{\scriptsize
\begin{center}
\begin{tabularx}{\linewidth}{p{3cm} X}
\toprule
Activation & Fast GELU approximation  \\
{} & \cite{hendrycks2016gelu} \\
Total batch size & 256 \\
Negatives per batch & 64 \\
Learning rate & 0.3 \\
Optimizer & Adadelta with $\rho=0.9$ \cite{Zeiler:2012ada} \\
ConveRT layers & Same as in \cite{Henderson:2019convert} \\
Input / template FFN layer size & 512 , 128 \\
Decoder FFN size & 256 \\
Decoder attention projections & 16 \\
\bottomrule
\end{tabularx}
\end{center}
}%
%}%
\vspace{-0.5mm}
\caption{ConVEx: Hyper-parameters at pretraining.}
\label{tab:hparams}
\vspace{-0.5mm}
\end{table}

\noindent \textbf{Pretraining: Technical Details.}
The ConVEx parameters at pretraining are randomly initialized, including the ConveRT layers, and the model is pretrained on the pairwise cloze Reddit data. Pretraining proceeds in batches of 256 examples, 64 of which are randomly paired sentences where no value should be extracted, and the remaining being pairs from the training data. This teaches the model that sometimes no value should be predicted, a scenario frequently encountered with slot labeling. Table~\ref{tab:hparams} provides a concise summary of these and other pretraining hyper-parameters.

\vspace{1.8mm}
\noindent \textbf{Computational Efficiency and Tractability.} ConVEx is pretrained for 18 hours on 12 Tesla K80 GPUs; this is typically sufficient to reach convergence. The total pretraining cost is roughly \$85 on Google Cloud Platform. This pretraining regime is orders of magnitude cheaper and more efficient than the prevalent pretrained NLP models such as BERT \cite{Devlin:2018arxiv}, GPT models \cite{Brown:2020gpt3}, XLNet \cite{Yang:2019neurips}, RoBERTa \cite{Liu:2019roberta}, etc. The reduced pretraining cost allows for wider experimentation, and aligns with recent ongoing initiatives on improving fairness and inclusion in NLP/ML research and practice \cite{Strubell:2019acl,Schwartz:2019green}.

\vspace{1.8mm}
\noindent \textbf{Fine-tuning: Technical Details.}
We use the same fine-tuning procedure for all fine-tuning experiments on all evaluation data sets. It proceeds for 4,000 steps of batches of size 64, stopping early if the loss drops below 0.001.\footnote{We enforce that exactly 20\% of examples in each batch contain a value, and 80\% contain no value. Further, the batch size is smaller than 64 in few-shot scenarios if the training set is too small to meet this ratio without introducing duplicates.} The ConveRT layers are frozen, while the other layers are initialized to their pretrained values and optimized with Adam \cite{adam:15}, with a learning rate of 0.001 that decays to $10^{-6}$ over the first 3,500 steps using cosine decay \cite{LoshchilovH17}. Dropout is applied to the output of the ConveRT layers with a rate of 0.5: it decays to 0 over 4,000 steps also using cosine decay. The residual layer for additional features (e.g., \textit{is\_requested}, \textit{token\_is\_numeric}) consists of a single 1024-dim hidden layer. As we demonstrate later in \S\ref{s:results}, this procedure works well across a variety of data settings. The early stopping and dropout are intended to prevent overfitting on very small data sets.

\vspace{1.8mm}
\noindent \textbf{Fine-tuning and Evaluation: Data and Setup.}
We rely on several diverse slot-labeling data sets, used as established benchmarks in previous work. First, we evaluate on a recent data set from \newcite{CoopeFarghly2020}: \textsc{restaurants-8k}, which comprises conversations from a commercial restaurant booking system. It covers 5 slots required for the booking task: \textit{date}, \textit{time}, \textit{people}, \textit{first name}, and \textit{last name}. Second, we use the Schema-Guided Dialog Dataset (SGDD) \cite{rastogi2019towards}, originally released for \textsc{dstc8}, in the same way as prior work \cite{CoopeFarghly2020}, extracting span annotated data sets from SGDD in four different domains. The particulars of the \textsc{restaurants-8k} and \textsc{dstc8} evaluation data are provided in the appendix.

Similar to \newcite{CoopeFarghly2020}, we simulate few-shot scenarios and measure performance on smaller sets sampled from the full data.  We (randomly) subsample the training sets of various sizes while maintaining the same test set.

%% In order to investigate the model behavior in low-data regimes, for both data sets 

Furthermore, we also evaluate ConVEx in the 5-shot evaluation task on the SNIPS data \cite{coucke2018snips}, following the exact setup of \newcite{hou2020few}, which covers 7 diverse domains, ranging from \textit{Weather} to \textit{Creative Work} (see Table~\ref{tab:snipsResults} later for the list of domains). The statistics of the SNIPS evaluation are also provided in the appendix.

The SNIPS evaluation task slightly differs from \textsc{restaurants-8k} and \textsc{dstc8}: we thus provide additional details related to fine-tuning and evaluation procedure on SNIPS, replicating the setup of \newcite{hou2020few}. Each of the 7 domains in turn acts as a held-out test domain, and the other 6 can be used for training. From the held-out test domain, episodes are generated that contain around 5 examples, covering all the slots in the domain. For each domain, we first further pretrain the ConVEx decoder layers (the ones that get fine-tuned) on the other 6 domains: we append the slot name to the template sentence, which allows training on all the slots. This gives a single updated fine-tuned ConVEx decoder model, trained on all slots of all other domains. For each episode, for each slot in the target domain we fine-tune 3 ConVEx decoders. The predictions are \textit{ensembled} by averaging probabilities to give final predictions. This helps reduce variability and improves prediction quality.

\vspace{1.8mm}
\noindent \textbf{Baseline Models.} For \textsc{restaurants-8k} and \textsc{dstc8}, we compare ConVEx to the current best-performing approaches from \newcite{CoopeFarghly2020}: Span-BERT and Span-ConveRT. Both models rely on the same CNN+CRF architecture\footnote{See \cite{CoopeFarghly2020} for further technical details.} applied on top of the subword representations transferred from a pretrained BERT(-Base/Large) model \cite{Devlin:2018arxiv} (Span-BERT), or from a pretrained ConveRT model \cite{Henderson:2019convert}.\footnote{\newcite{CoopeFarghly2020} also evaluated an approach based on the same CNN+CRF architecture as Span-\{BERT, ConveRT\} which does not rely on any pretrained sentence encoder, and learns task-specific subword representations from scratch. However, that approach is consistently outperformed by Span-ConveRT, and we therefore do not report it for brevity.} Similar to \newcite{CoopeFarghly2020}, for each baseline we run hyper-parameter optimization via grid search, evaluating on the dev set of \textsc{restaurants-8k}.

For SNIPS, we compare ConVEx to a wide spectrum of different few-shot learning models proposed and compared by \newcite{hou2020few}.\footnote{A full description of each baseline model is beyond the scope of this work, and we refer to \cite{hou2020few} for further details. For completeness, short summaries of each baseline model on SNIPS are provided in the appendix.}

%% (IV, this has been removed)
%%Similar to \newcite{CoopeFarghly2020}, for each baseline we run hyper-parameter optimization via grid search, evaluating on the dev set of \textsc{restaurants-8k}.

%%\ivulic{How did you optimise hparams? Is this correct? How large is the dev set? \matt{there is no hparams grid search, there wasn't one for the spanconvert paper either. rather each approach (spanconvert, uve, spanbert) has a single set of hparams used across all domains, slots, and dataset size. those are each optimized by hand, and we weren't strict about optimizing these on a separate dev set... though there is little room for overfitting} }

One crucial difference between our approach and the methods evaluated by \newcite{hou2020few} is as follows: we treat each slot independently, using separate ConVEx decoders for each, while the their methods train a single CRF decoder that models all slots jointly. One model per slot is simpler, easier for practical use (e.g., it is possible to keep and manage data sets for each slot independently), and makes pretraining conceptually easier.\footnote{Moreover, the methods of \newcite{hou2020few} are arguably more computationally complex: at inference, their strongest models (i.e., TapNet and WPZ, see the appendix, run BERT for every sentence in the fine-tuning set (TapNet), or run classification for every pair of test words and words from the fine-tuning set (WPZ). The computational complexity of the ConVEx approach does not scale with the fine-tuning set, only with the number of words in the query sequence.}

\vspace{1.8mm}
\noindent \textbf{Evaluation Measure.}
Following previous work \cite{coucke2018snips,rastogi2019towards,CoopeFarghly2020}, we report the average $F_1$ scores for extracting the correct span per user utterance. If the models extract part of the span or a longer span, this is treated as an incorrect span prediction.

%%they also learn a joint model over all slots, while UVE learns separate models. 

\section{Results and Discussion}
\label{s:results}

\begin{figure*}[!t]
    \centering
    \begin{subfigure}[!t]{0.478\linewidth}
        \centering
        \includegraphics[width=0.88\linewidth]{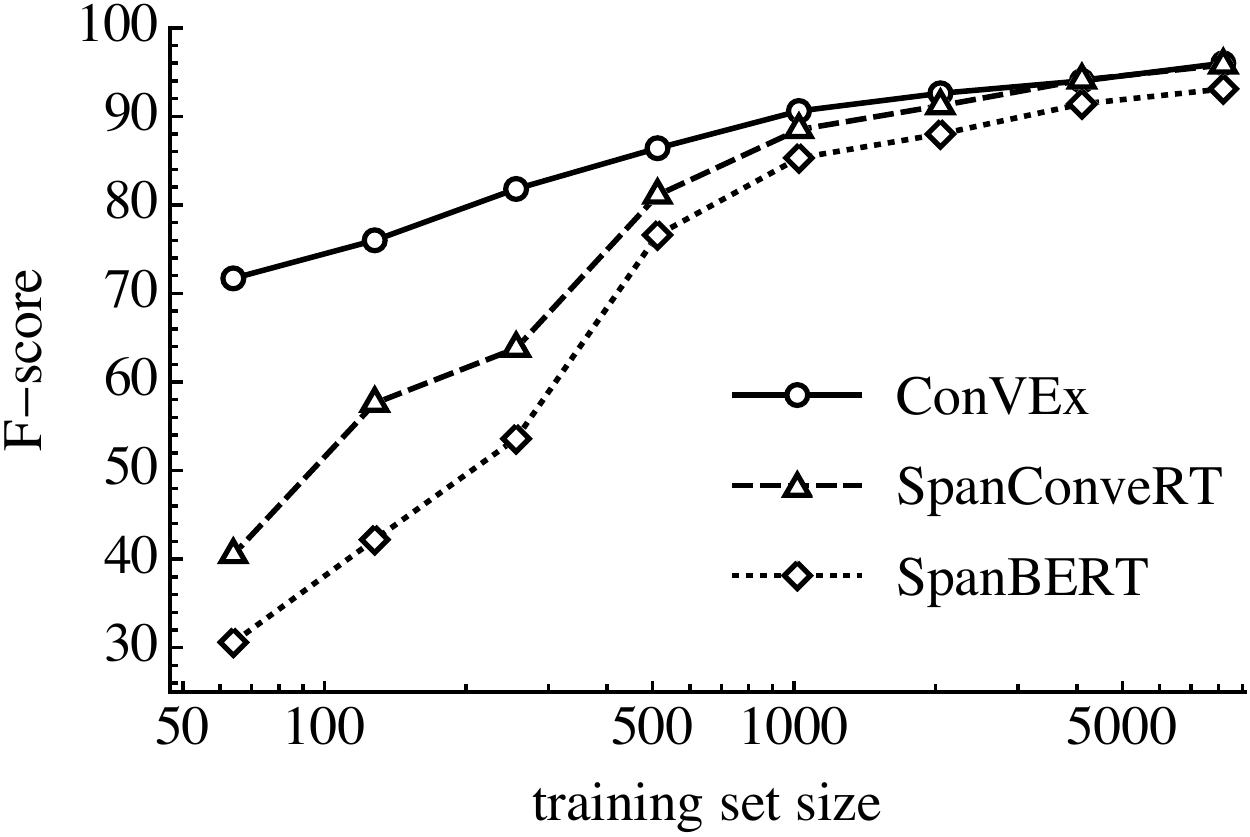}
        %\vspace{-0.7em}
        \caption{\textsc{restaurants-8k}}
        \label{fig:restaurantsPlot}
    \end{subfigure}
    \begin{subfigure}[!t]{0.478\textwidth}
        \centering
        \includegraphics[width=0.88\linewidth]{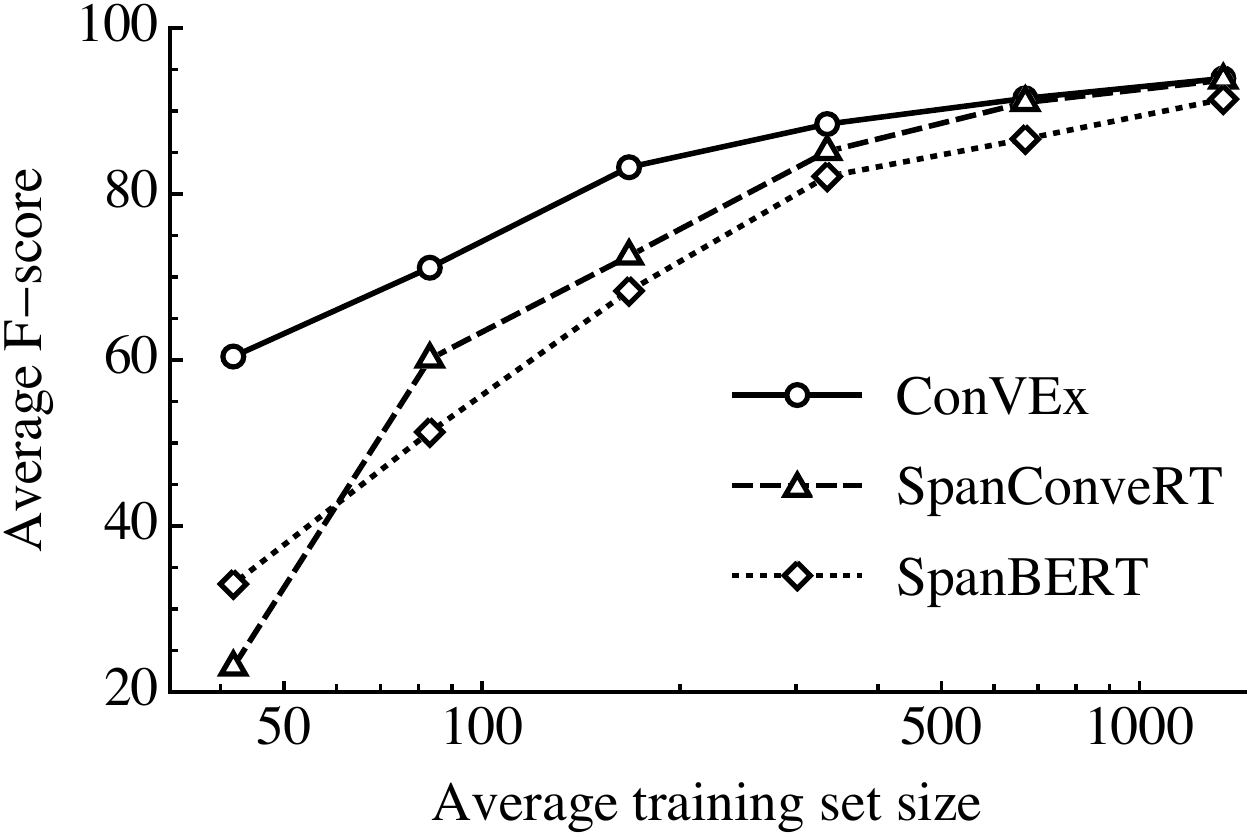}
        %\vspace{-0.7em}
        \caption{\textsc{dstc8}}
        \label{fig:dstc8Plot}
    \end{subfigure}
    \vspace{-0.5mm}
    \caption{Average $F_1$ across all slots for (a) \textsc{restaurants-8k}, and (b) \textsc{dstc8}, with varying training set sizes.}
    \vspace{-0.5mm}
\label{fig:plots}
\end{figure*}

%% IV: For camera-ready
%%\matt{we can add results for pretraining on reddit vs cc100 vs wikipedia}

\noindent \textbf{Intrinsic (Reddit) Evaluation.} 
ConVEx reaches a precision of 84.8\% and a recall of 85.3\% on the held-out Reddit test set (see Table~\ref{tab:redditData} again), using 25\% random negatives as during pretraining. The ConVEx variant without the auxiliary loss (termed \textit{no-aux} henceforth) reaches a precision of 82.7\% and a recall of 83.9\%, already indicating the usefulness of the auxiliary loss.\footnote{While we evaluate the two ConVEx variants also in the slot-labeling tasks later, unless noted otherwise, in all experiments we assume the use of the variant \textit{with} the \textit{aux} loss.} These preliminary results serve mostly as a sanity check, suggesting ConVEx's ability to generalize over unseen Reddit data; we now evaluate its downstream task efficacy. %in the subsequent experiments.

\begin{table*}[t]
\begin{center}
\def\arraystretch{1.05}
\resizebox{0.99\textwidth}{!}
{ \footnotesize
    \begin{tabularx}{\textwidth}{l *{7}{Y} Y}
        \toprule
          &
          {\bf Weather} &
          {\bf Music} &
          {\bf Playlist} &
          {\bf Book} &
          {\bf Search} &
          {\bf Restaurant} &
          {\bf Creative} &
          {\it Average }\\ \cmidrule{2-9}
          
          \hspace{-3mm}\textit{\newcite{hou2020few}} & & & & & & & & \\ 
          TransferBERT & 59.4 & 42.0 & 46.1 & 20.7 & 28.2 & 67.8 & 58.6 & 46.1 \\ 
          
          SimBERT & {53.5} & {54.1} & {42.8} & {75.5} & {57.1} & {55.3} & {32.4} & 52.9   \\
          
          WPZ+BERT & {67.8} & {56.0} & {46.0} & {72.2} & {73.6} & {60.2} & {66.9} & 63.2\\
          
          TapNet & {53.0} & {49.8} & {54.9} & {83.4} & {63.1} & {59.8} & {67.0} & 61.6 \\
          
          TapNet+CDT & {66.5} & {66.4} & {68.2} & \textbf{{85.8}} & {73.6} & {64.2} & {68.5} & 70.4 \\
          
          L-WPZ+CDT & \textbf{{74.7}} & {56.7} & {52.2} & {78.8} & {80.6} & {69.6} & {67.5} & 68.6 \\
          
          L-TapNet+CDT & {71.6} & {67.2} & {{75.9}} & {84.4} & {{82.6}} & {70.1} & \textbf{{73.4}} & 75.0\\ %\midrule
          
          \hspace{-3mm}\textit{This work} & & & & & & & & \\
          
          ConVEx (with aux) & {71.5} & \textbf{{77.6}} & \textbf{{79.0}} & {{84.5}} & \textbf{{84.0}} & \textbf{{73.8}} & {67.4} & \textbf{76.8} \\ \bottomrule
    \end{tabularx}
}
\end{center}
\vspace{-0.5mm}
\caption{$F_1$ scores on SNIPS 5-shot evaluation, following the exact setup of \newcite{hou2020few}. For an overview of the baseline models from \newcite{hou2020few}, see the original work and short summaries available in the appendix.}
\label{tab:snipsResults}
\vspace{-0.5mm}
\end{table*}
\begin{table}[t]
\def\arraystretch{0.99}
\resizebox{0.49\textwidth}{!}
{
\begin{tabularx}{0.6\textwidth}{r *{2}{Y}}
\toprule
\textbf{Fraction} & \textbf{ConVEx} & \textbf{ConVEx ensemble} \\ \cmidrule(lr){2-3}
1 (8198)        & \textbf{96.0} & 95.8 \\
$\nicefrac{1}{2}$ (4099)     & {94.1} & \textbf{94.2}  \\
$\nicefrac{1}{4}$ (2049)     & \textbf{92.5}  & \textbf{92.5}    \\
$\nicefrac{1}{8}$ (1024)     & {90.6}  & \textbf{90.7}   \\
$\nicefrac{1}{16}$ (512)     & {86.4}  & \textbf{88.2}     \\
$\nicefrac{1}{32}$ (256)     & {81.8}  & \textbf{83.9}   \\
$\nicefrac{1}{64}$ (128)     & {76.0}  & \textbf{78.2}     \\
$\nicefrac{1}{128}$ (64)    & {71.7}    & \textbf{73.5}      \\
\bottomrule
\end{tabularx}
}
\vspace{-0.5mm}
\caption{Average $F_1$ scores across all slots for \textsc{restaurants-8k} for ConVEx with and without ensembling. The ConVEx ensemble model fine-tunes 3 decoders per slot, and then averages their output scores.}
\label{tab:ensembleResults}
\vspace{-0.5mm}
\end{table}

\vspace{1.8mm}
\noindent \textbf{Evaluation on \textsc{restaurants-8k} and \textsc{dstc8}.} The main respective results are summarized in Figure~\ref{fig:restaurantsPlot} and Figure~\ref{fig:dstc8Plot}, with additional results available in the appendix. In full-data scenarios all models in our comparison, including the baselines from \newcite{CoopeFarghly2020}, yield strong performance reaching $\geq 90\%$ or even $\geq 95\%$ average $F_1$ across the board.\footnote{The only exception is Span-BERT's lower performance on the \textsc{dstc8} \textit{Homes\_1} evaluation, see the appendix. In general, as shown previously by \newcite{CoopeFarghly2020} and revalidated here, conversational pretraining based on response selection (ConveRT) seems more useful for conversational applications than regular LM-based pretraining (BERT).} However, it is encouraging that ConVEx is able to surpass the baseline models on average even in the full-data regimes. 

%%  we additionally plot the performance of ConVEx along with the baseline models in few-shot scenarios with varying numbers of examples

%% IV: Add this to Camera-ready
%% \matt{we should add results from \newcite{namazifar2020}}

Figure~\ref{fig:restaurantsPlot} and Figure~\ref{fig:dstc8Plot} also suggest true benefits of the proposed ConVEx approach: the ability of ConVEx to handle few-shot scenarios well. The gap between ConVEx and the baseline models becomes more and more pronounced as we continue to reduce the number of annotated examples for the labeling task. On \textsc{restaurants-8k} the gain is still small when dealing with 1,024 annotated examples (+2.1 $F_1$ points over the strongest baseline), but it increases to +18.4 $F_1$ points when 128 annotated examples are available, and further to +31.2 $F_1$ points when only 64 annotated examples are available. We can trace a similar behavior on \textsc{dstc8}, with gains reported for all the \textsc{dstc8} single-domain subsets in few-shot setups.

These results point to the following key conclusion. While pretrained representations are clearly useful for slot-labeling dialog tasks, and the importance of pretraining becomes increasingly important when we deal with few-shot scenarios, the chosen pretraining paradigm has a profound impact on the final performance. The pairwise cloze pretraining task, tailored for slot-labeling tasks in particular, is more robust and better adapted to few-shot slot-labeling tasks. This also verifies our hypothesis that it is possible to learn effective domain-specific slot-labeling systems by simply fine-tuning a pretrained general-purpose slot labeler relying only on a handful of domain-specific examples.

\vspace{1.8mm}
\noindent \textbf{SNIPS Evaluation (5-Shot).} The versatility of ConVEx is further verified in the 5-shot labeling task on SNIPS following \newcite{hou2020few}'s setup. The results are provided in Table~\ref{tab:snipsResults}. We report the highest average $F_1$ scores with ConVEx; ConVEx also surpasses all the baselines in 4/7 domains, while the highest scores in the remaining three domains are achieved by three different models from \newcite{hou2020few}. This again hints at the robustness of ConVEx, especially in few-shot setups, and shows that a single pretrained model can be adapted to a spectrum of slot-labeling tasks and domains.

%% on SNIPS as well as on the other two data sets 
These results also stand in contrast with the previous findings of \newcite{hou2020few} where they claimed ``\textit{...that fine-tuning on extremely limited examples leads to poor generalization ability}''. On the contrary, our results validate that it is possible to fine-tune a pretrained slot-labeling model directly with a limited number of annotated examples for various domains, without hurting the generalization ability of ConVEx. In other words, we demonstrate that the mainstream ``pretrain then fine-tune'' paradigm is a viable solution to sequence-labeling tasks in few-shot scenarios, but with the condition that the pretraining task must be structurally well-aligned with the intended downstream tasks.

\vspace{1.8mm} 
Next, we analyze the benefits of model ensembling, as done in the 5-shot SNIPS task, also on \textsc{restaurants-8k}. The results across different training data sizes are shown in Table~\ref{tab:ensembleResults}. While there is no performance difference when a sufficient number of annotated examples is available, the scores suggest that the model ensembling strategy does yield small but consistent improvements in few-shot scenarios, as it mitigates the increased variance that is typically met in these setups.

\begin{figure}[t!]
    \centering
    \includegraphics[width=1.0\linewidth]{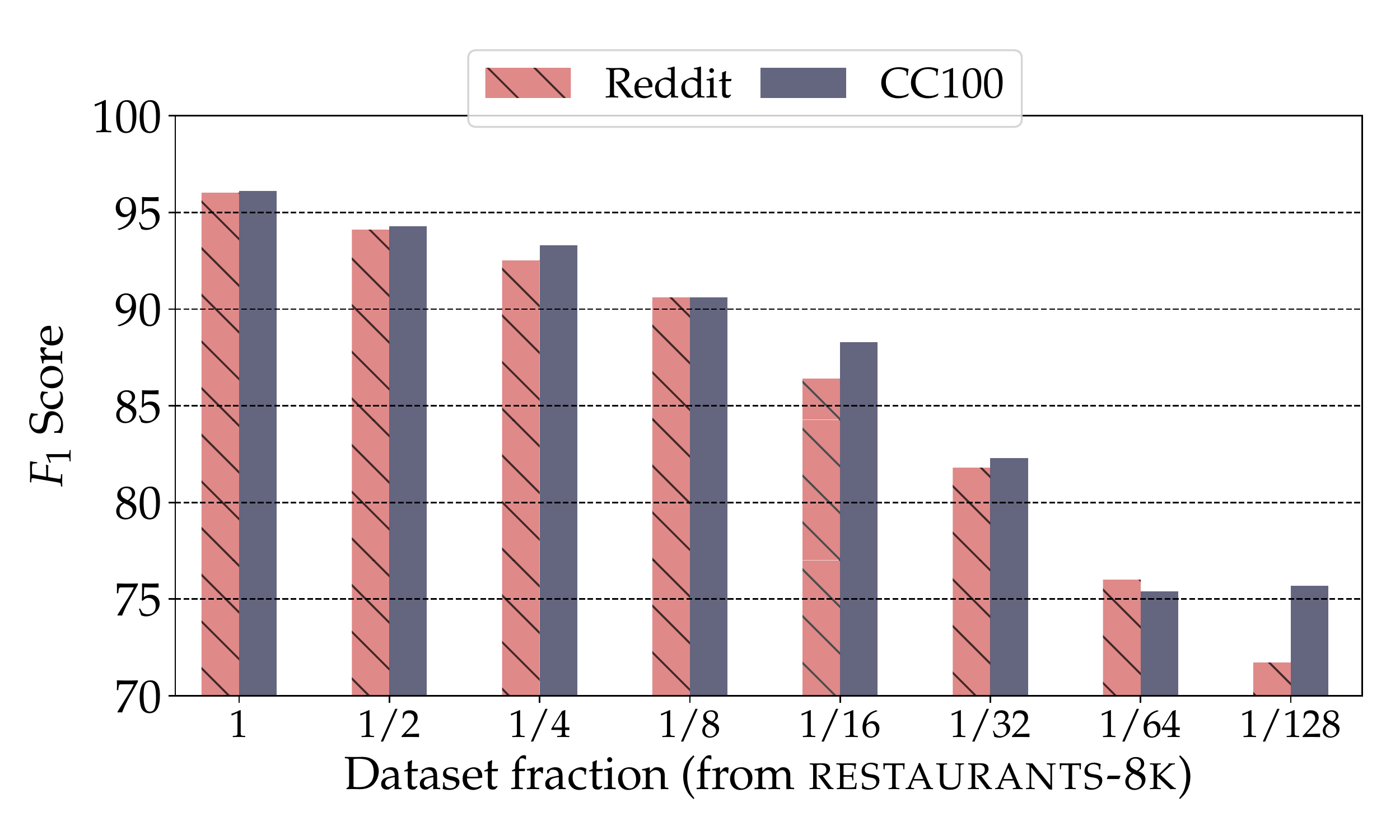}
    %\vspace{-2.5mm}
    \caption{$F_1$ scores on \textsc{restaurants-8k} (averaged over all slots) with varying training data sizes when ConVEx is pretrained on Reddit versus CC100.}
    \label{fig:cc100}
    \vspace{-0.5mm}
\end{figure}

\vspace{1.8mm}
\noindent \textbf{Pretraining on CC100.} 
We also test the robustness of ConVEx by pretraining it on another large Web-scale dataset: CC100 \cite{Wenzek:2020lrec,Conneau:2020acl} is a large CommonCrawl corpus available for English and more than 100 other languages. We use the English CC100 portion to pretrain ConVEx relying on exactly the same procedure described in \S\ref{s:methodology}, and then fine-tune it as before. First, its intrinsic evaluation on the held-out test set already hints that the CC100-based ConVEx is also a powerful slot labeller: we reach a precision of 85.9\% and recall of 86.3\%. More importantly, the results on \textsc{restaurants8k}, provided in Figure~\ref{fig:cc100}, confirm that another general-purpose corpus can be successfully used to pretrain the ConVEx model. We even observe slight gains on average over the Reddit-based model.

\begin{figure}[t!]
    \centering
    \includegraphics[width=1.0\linewidth]{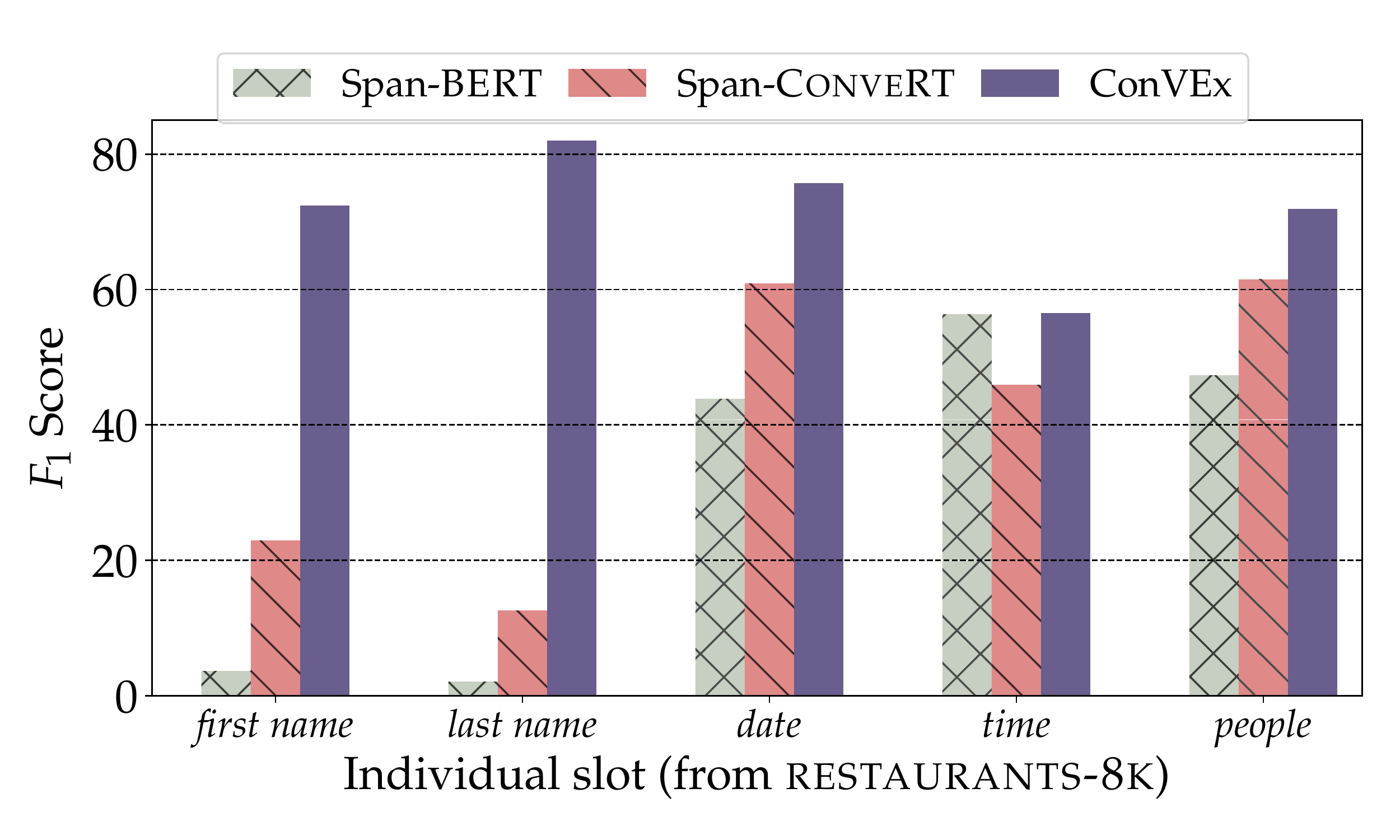}
    %\vspace{-2.5mm}
    \caption{Average $F_1$ scores for each slot for the Span-ConveRT and ConVEx models trained on $\nicefrac{1}{128}$ of the  \textsc{restaurants-8k} training set, i.e., 64 examples.}
    \label{fig:perslot}
    \vspace{-0.5mm}
\end{figure}
\begin{figure}[t!]
    \centering
    \includegraphics[width=1.0\linewidth]{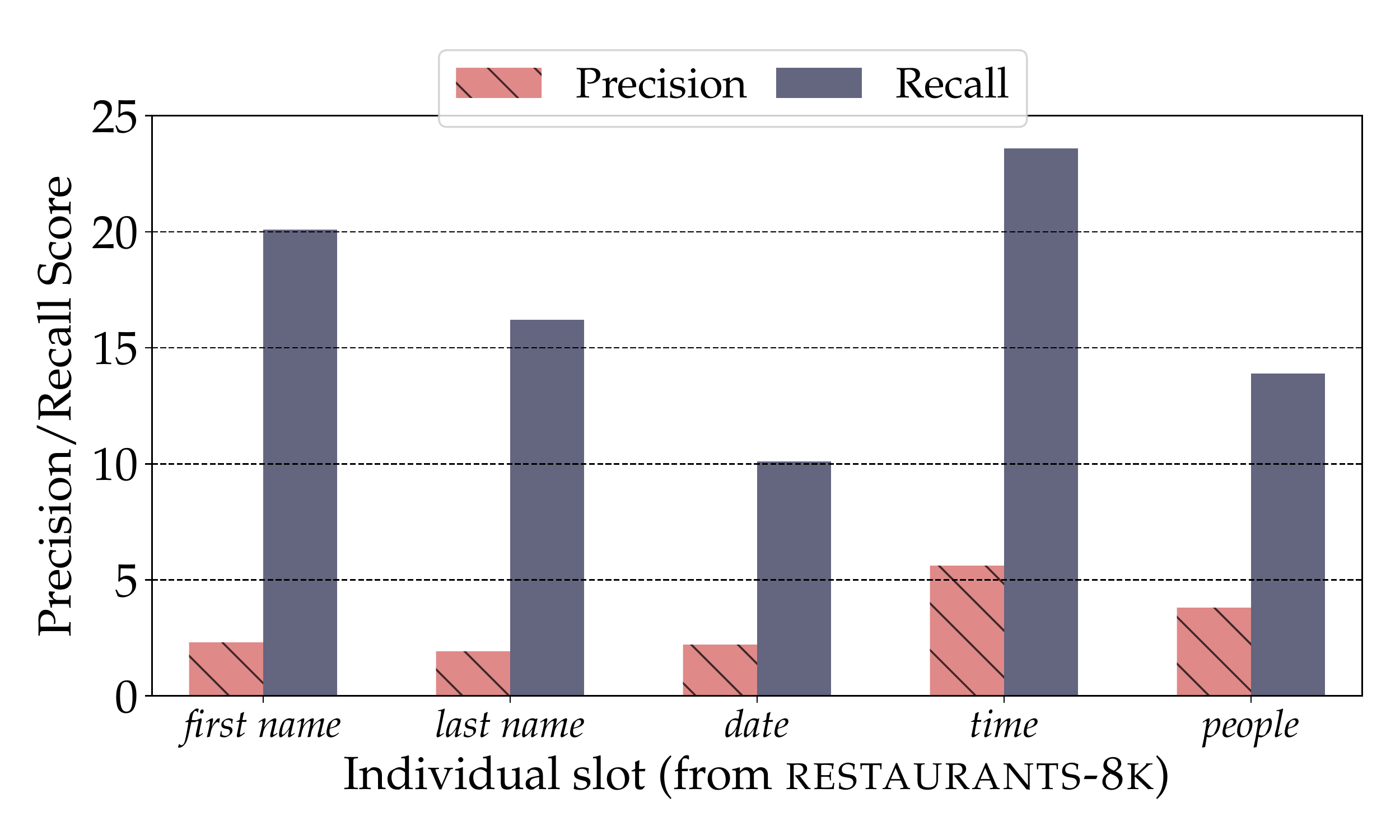}
    %\vspace{-2.5mm}
    \caption{Performance of the ConVEx decoder across all slots in \textsc{restaurants-8k} without any fine-tuning.}
    \label{fig:no-finetuning}
    \vspace{-0.5mm}
\end{figure}

\vspace{1.8mm}
\noindent \textbf{Inductive Bias of ConVEx.} 
In sum, ConVEx outperforms current state-of-the-art slot-labeling models such as Span-ConveRT, especially in low-data settings, where the performance difference is particularly large. The model architectures of Span-\{BERT, ConveRT\} and ConVEx are very similar: the difference in performance thus arises mainly from the pretraining task, and the fact that ConVEx's sequence-decoding layers are pretrained, rather than learned from scratch. We now analyse the inductive biases of ConVEx, that is, how the pretraining regime and the main assumptions affect its behavior before and after fine-tuning. 

%that is, the behaviors that UVE models are biased towards with little fine-tuning.

%% as the previous state-of-the-art models on this evaluation set.

First, we analyze \textit{per-slot performance} on \textsc{restaurants-8k}, comparing ConVEx (\textit{with aux}) with Span-BERT and Span-ConveRT. The scores in a few-shot scenario with 64 examples are provided in Figure~\ref{fig:perslot}, and we observe similar patterns in other few-shot scenarios. The results indicate the largest performance gap for the slots \textit{first name} and \textit{last name}. This is expected, given that by the ConVEx design the keyphrases extracted from Reddit consist of rare words, and are thus likely to cover plenty of names without sufficient coverage in small domain-specific data sets. Nonetheless, we also mark prominent gains over the baselines achieved also for the other slots with narrower semantic fields, where less lexical variability is expected (\textit{date} and \textit{people}).

%Table~\ref{tab:restaurants128} presents a slot-level breakdown of the \textsc{restaurants-8k} evaluation in the smallest training set condition. 

%% (IV, said before)
%% (even though the user input contains no \emph{BLANK})

We can also expose ConVEx's built-in biases by applying it with no fine-tuning. Figure~\ref{fig:no-finetuning} shows the results with no slot-specific fine-tuning on \textsc{restaurants-8k}, feeding the user input as both the template and input sentence. We extract at most one value from each sentence, where the model predicted a value for 96\% of all the test examples, 16\% of which corresponded to an actual labeled slot, and 86\% did not. The highest recalls were for the name slots, and the time slot, which correlates with the slot-level breakdown results from Figure~\ref{fig:perslot}.\footnote{The most frequent predictions from non-finetuned ConVEx that do not correspond to a labeled slot on \textsc{restaurants-8k} give further insight into its inductive biases. The top 10 extracted non-labeled values are in descending order: \textit{booking, book, reservation, a reservation, a table, indoors, restaurant, cuisine, outside table,} and \textit{outdoors}. Some of these could be modeled as slot values with an extended ontology, such as \textit{indoors} or \textit{outdoors/outside table}.}

\iffalse
\begin{table}[htbp]
\def\arraystretch{0.91}
\resizebox{0.47\textwidth}{!}
{
\begin{tabularx}{0.6\textwidth}{r *{2}{Y}}
\toprule
\textbf{Slot} & \textbf{Precision} & \textbf{Recall} \\ \cmidrule(lr){2-3}
first name & 2.3 & 20.1 \\
last name & 1.9 & 16.2 \\
date & 2.2 & 10.1 \\
time & 5.6 & 23.6 \\
people & 3.8 & 13.9 \\
\bottomrule
\end{tabularx}
}
\caption{Precision and Recall for all slots in \textsc{restaurants-8k} for the ConVEx decoder before any fine-tuning. This illustrates the bias inherent in the model.}
\label{tab:noFinetune}
\vspace{-1mm}
\end{table}
\fi

%% IV: with 'vegetarian' there were 11 values...

%\section{Conclusion and Future Work}
\section{Conclusion}
\label{s:conclusion}

%% Reformulated from the abstract
We have introduced \textit{ConVEx} (Conversational Value Extractor), a light-weight pretraining and fine-tuning neural approach to slot-labeling dialog tasks. We have demonstrated that it is possible to learn domain-specific slot labelers even in low-data regimes by simply fine-tuning decoder layers of the pretrained general-purpose ConVEx model. The ConVEx framework has achieved a new leap in performance on standard dialog slot-labeling tasks, most notably in few-shot setups, by aligning the pretraining phase with the downstream fine-tuning phase for slot-labeling tasks.

%The ConVEx framework achieves a new leap in performance by aligning the pretraining phase with the downstream fine-tuning phase for sequence labeling tasks. This is realized through the novel \textit{pairwise cloze} pretraining objective, a sentence-pair value extraction task that is structurally aligned with the downstream slot-labeling tasks. While existing approaches typically require learning additional sequence level layers from scratch, ConVEx requires no new layers and can be fully fine-tuned. We have validated the effectiveness and usefulness of the ConVEx approach to slot labeling across a spectrum of diverse slot-labeling domains and data sets, reporting state-of-the-performance in full-data setups, as well as the strongest gains in the most challenging, few-shot setups. 

%% IV: Add future work and expand this in the CR
In future work, we plan to investigate the limits of data-efficient slot labeling, focusing on one-shot and zero-shot setups. We will also apply ConVEx to related tasks such as named entity recognition and conversational question answering.

%%future work could include the one-shot setting, representing a slot a slot with a single template sentence with no fine-tuning- or other approaches that use the UVE as is, passing in a template sentence with a BLANK, rather than forget about BLANKs and finetune.
%future work may also apply UVE to related tasks- NER, conversational QA?
%\ivan{IV: write tomorrow}

%acknowledgements? Thanks to Yutai Hou for sharing the data and evaluation episodes for SNIPS. thanks to colleagues at PolyAI for discussions etc. 

%\clearpage
\section*{Ethical Considerations}
To the best of our knowledge, the conducted work does not imply any undesirable ethical ramifications. By design and its uncontrollable nature, the Reddit data does encode a variety of societal, gender, and other biases; however, the models pretrained on the Reddit data are always fine-tuned for specific tasks using controlled data, and the Reddit-pretrained models are not used for any text generation nor full-fledged dialogue applications directly. The evaluation data used in this work have been collected in previous work following standard crowdsourcing and data annotation practices.

%% IV: Return to camera-ready
\section*{Acknowledgements}
We would like to thank Yutai Hou for sharing the data and evaluation episodes for the SNIPS evaluation. Thanks to our colleagues at PolyAI for fruitful discussions and critical examinations of this work. We would also like to thank Sam Coope and Tyler Farghly for their help with rerunning and validating Span-BERT and Span-ConveRT.

%\clearpage
\bibliography{refs, other_refs}
\bibliographystyle{acl_natbib}

%\iffalse
\clearpage
\label{s:appendix}
\appendix
\twocolumn
\section{Evaluation Data Statistics}
\label{ss:evalStats}

For completeness, we provide the summary stats of the evaluation data used in our work:
\begin{description}[style=unboxed,leftmargin=0cm]
        \item [Table~\ref{tab:r8k-data}] shows the statistics of the \textsc{restaurants-8k} data set. The data set is available at: \\ {\small \url{github.com/PolyAI-LDN/task-specific-datasets}}.
    \item [Table~\ref{tab:dstc8-data}] shows the statistics of the \textsc{dstc8} data set. The data set is available at: \\ {\small \url{github.com/PolyAI-LDN/task-specific-datasets}}.
    \item [Table~\ref{tab:snips-full}] provides the statistics of the original SNIPS data set \cite{coucke2018snips}, For further details on how the data set has been used in the 5-shot evaluation setup we refer the reader to the work of \newcite{hou2020few}. The data sets are available at: \\ \url{github.com/AtmaHou/FewShotTagging}
\end{description}

\noindent Recently, \textsc{restaurants-8k} and \textsc{dstc8} training and evaluation data have been made available via the integrated DialoGLUE benchmark \cite{Mehri:2020dialoglue}. For further details regarding the two evaluation sets, we also refer the reader to the original work \cite{rastogi2019towards,CoopeFarghly2020}.

\onecolumn
\begin{table}[!t]
\centering
%\resizebox{0.75\textwidth}{!}{%
{\footnotesize
\begin{tabularx}{0.8\textwidth}{r YYYYYY} \toprule
      & \textit{people}   & \textit{time}     & \textit{date}     & \textit{first name} & \textit{last name} & Total \\ \midrule
train & \textbf{2164} (547) & \textbf{2164} (547) & \textbf{1721} (601) &  \textbf{887} (364)    & \textbf{891} (353)    & 8198  \\
test   & \textbf{983} (244)  & \textbf{853} (276)  & \textbf{802} (300)  &  \textbf{413} (177)    & \textbf{426} (174)    & 3731
\\ \bottomrule
\end{tabularx}
}%
%}%
%\vspace{-1mm}
\caption{The number of examples for each slot in the \textsc{restaurants-8k} data set. Numbers in brackets show how many examples have the slot \textit{is\_requested}.}
\label{tab:r8k-data}
%\vspace{-1mm}
\end{table}

\begin{table*}[t]
\begin{center}
{\small
\renewcommand*{\arraystretch}{1.5}
\begin{tabularx}{0.8\textwidth}{r Y Y p{60mm}}

\toprule
   \textbf{Sub-domain} & \textbf{Train Size} & \textbf{Test Size} & \textbf{Slots}                                                                                                                     \\ \midrule
\textbf{Buses\_1}      & 1133                & 377               & 
from\_location (169/54), leaving\_date (165/57),
to\_location (166/52)                  \\ 
\textbf{Events\_1}     & 1498                & 521               & 
city\_of\_event (253/82), date (151/33), 
subcategory (56/26)                               \\ 
\textbf{Homes\_1}      & 2064                & 587               & area (288/86), visit\_date (237/62)                                                                                                  \\ 
\textbf{RentalCars\_1} & 874                 & 328               & 
dropoff\_date (112/42), pickup\_city (116/48),  pickup\_date (120/43), pickup\_time (119/43)\\
\bottomrule
\end{tabularx}
}
\end{center}
\caption{Statistics of the used data sets extracted from the \textsc{dstc8} schema-guided dialog dataset. The number of examples in the train and test sets for each slot are reported in parentheses.}
\label{tab:dstc8-data}
%\vspace{-5mm}
\end{table*}

\begin{table}[!t]
\begin{center}
{\small
\begin{tabularx}{80mm}{r Y Y}
\toprule
   \textbf{Domain} & \textbf{\# of Sentences} & \textbf{Labels} \\    
   \cmidrule(lr){2-3}
   {Weather} & {2,100} & {10} \\
   {Music} & {2,100} & {10} \\
   {Playlist} & {2,042} & {6} \\
   {Book} & {2,056} & {8} \\
   {Search} & {2,059} & {8} \\
   {Restaurants} & {2,073} & {15} \\
   {Creative} & {2,054} & {3} \\
\bottomrule
\end{tabularx}
}%
\end{center}
\caption{Statistics of the original SNIPS data set.}
\label{tab:snips-full}
%\vspace{-5mm}
\end{table}

\twocolumn
\section{Baseline Models in the SNIPS Evaluation}
\label{ss:snipsBaselines}

This appendix provides a brief summary of the models from \newcite{hou2020few} included in the SNIPS evaluation (Table~\ref{tab:snipsResults}) alongside ConVEx.

\begin{description}[style=unboxed,leftmargin=0cm]
    \item [TransferBERT] is a direct application of BERT \cite{Devlin:2018arxiv} to sequence labeling. It is first trained on the source domains. As the sequence labeling layers are domain-specific, these are then removed, and new layers are fine-tuned on the in-domain training set (i.e., \newcite{hou2020few} refer to it as the \textit{support set}; this is exactly what we use for fine-tuning ConVEx).
    \item [SimBERT] predicts sequence labels according to the cosine similarity between the representations from a BERT model of the input tokens with tokens in the support set, selecting the labels of the most similar labeled tokens.
    \item [WarmProtoZero (WPZ)] \cite{Fritzler2019} applies Prototypical Networks \cite{snell2017prototypical} to sequence labeling tasks. It treats sequence-labeling as word-level classification, and can either use randomly initialized word embeddings, or pretrained representations in the case of WPZ+BERT.
    \item [TapNet] is a few-shot learning paradigm originally applied to image classification \cite{yoon2019tapnet}. This works similarly to Prototypical Networks, but includes a task-adaptive network that projects examples into a space where words of differing labels are well separated.
    \item [Collapsed Dependency Transfer (CDT)] is a technique for simplifying  transition dynamics of a CRF, applied to both TapNet and WPZ. This represents the full transition matrix using shared abstract transitions, e.g. modeling transitions between any \emph{Begin} tag to the \emph{Begin} tag of any different slot using a shared probability. 
    \item[Label Enhanced] models, denoted \emph{L-WPZ} and \emph{L-TapNet} use the semantics of the label names themselves to enrich the word-label similarity modeling. 
\end{description}

\section{Additional Results}
The exact $F_1$ scores corresponding to the results plotted in Figure~\ref{fig:restaurantsPlot} and Figure~\ref{fig:dstc8Plot} are provided in Table~\ref{tab:restaurantsResults} and Table~\ref{tab:dstc8Results}, respectively. Additional results with model ensembling are available in Table~\ref{tab:ensembleResultsFull}.

\onecolumn

\begin{table}[t]
\centering
\def\arraystretch{1.0}
{\footnotesize
%\resizebox{0.\textwidth}{!}
%{%
\begin{tabularx}{0.8\textwidth}{r *{4}{Y}}
\toprule
\textbf{Fraction} & \textbf{Span-ConveRT}  & \textbf{Span-BERT} & \textbf{ConVEx -no-aux} & \textbf{ConVEx -full} \\ \midrule
1 (8198)        & {95.8}              & 93.1  & 95.6   & \textbf{96.0} \\
$\nicefrac{1}{2}$ (4099)     & \textbf{94.1}                & 91.4 & \textbf{94.1}      & \textbf{94.1} \\
$\nicefrac{1}{4}$ (2049)     & {91.2}              & 88.0 & 92.2 & \textbf{92.6}      \\
$\nicefrac{1}{8}$ (1024)     &{88.5}             & 85.3  & 90.0 & \textbf{90.6}     \\
$\nicefrac{1}{16}$ (512)     & {81.1}            & 76.6 & 86.2 &  \textbf{86.4}       \\
$\nicefrac{1}{32}$ (256)     &{63.8}               & 53.6 & 78.4 & \textbf{81.8}     \\
$\nicefrac{1}{64}$ (128)     & {57.6}              & 42.2& 73.4 & \textbf{76.0}       \\
$\nicefrac{1}{128}$ (64)    & {40.5}              & 30.6 & 70.9 &  \textbf{71.7}      \\
\bottomrule
\end{tabularx}
%}
}%
%\vspace{-1mm}
\caption{Average $F_1$ scores across all slots for the evaluation on \textsc{restaurants-8k} test data with varying training set fractions. Numbers in brackets denote the training set sizes. The peak scores in each training setup (i.e., per row) are in bold. }
\label{tab:restaurantsResults}
\vspace{-1mm}
\end{table}
\begin{table}[!t]
\centering
\def\arraystretch{1.0}
{\footnotesize
%\resizebox{0.8\textwidth}{!}
%{%
\begin{tabularx}{0.9\textwidth}{p{4mm} *{5}{Y}}
\toprule

       & \textbf{Fraction} & \textbf{Span-ConveRT} & \textbf{Span-BERT} & \textbf{ConVEx -no-aux} &  \textbf{ConVEx} \\ \midrule
Buses\_1  \\
& 1 (1133)       & {93.5}                 & 93.3    &95.1  & \textbf{96.0} \\
              & \nicefrac{1}{2} (566)      & {88.9}               & 85.3  &90.4& \textbf{92.6}    \\
              & \nicefrac{1}{4} (283)     & {84.0}                 & 77.8  & \textbf{88.6} & {86.7}    \\
              & \nicefrac{1}{8} (141)      & {69.1}                & 69.6 & 83.6  & \textbf{84.0}     \\
              & \nicefrac{1}{16} (70)      & {58.3}                & 44.4 & 74.5 & \textbf{75.2}     \\ 
              & \nicefrac{1}{32} (35) & 32.7 &25.5  & \textbf{65.4} & 59.2\\ \midrule
Events\_1 \\
& 1 (1498)       & \textbf{92.7}                 & 84.3  & {92.4} & 91.7  \\
              & \nicefrac{1}{2} (749)     & {86.9}                & 80.2 & \textbf{88.4} & {87.3}     \\
              & \nicefrac{1}{4} (374)      & {82.2}                & 78.6  & 86.4 &  \textbf{87.2}     \\
              & \nicefrac{1}{8} (187)      & {70.0}             & 57.4   & 72.4& \textbf{82.2}    \\
              & \nicefrac{1}{16} (93)     & {55.9}               & 43.9  & 65.7 & \textbf{66.6}    \\
              & \nicefrac{1}{32} (47) & 39.2 & 25.6 & 51.4 & \textbf{54.0}\\ \midrule
Homes\_1  \\
              & 1 (2064)                    &{ 94.8 }             & {96.3}  & 95.5 & \textbf{98.3}    \\
              & \nicefrac{1}{2} (1032)      & \textbf{96.1}                & \textbf{95.7} & 95.6 & {95.6}      \\
              &\nicefrac{1}{4} (516)     & \textbf{95.4}                & \textbf{95.1} & 93.0 &  {94.5}      \\
              & \nicefrac{1}{8} (258)       & {93.4}                & 89.5   & 92.2 & \textbf{94.8}    \\
              &\nicefrac{1}{16} (129)     & {86.3}                & 76.4  & \textbf{94.0}  & {92.3}   \\ 
              & \nicefrac{1}{32} (65) & 77.1 & 61.2 & 89.4 & \textbf{92.0} \\ \midrule
RentalCars\_1 \\
& 1 (874)                                  & \textbf{94.0}              & 92.8 & 90.7& {92.0}      \\
              & \nicefrac{1}{2} (437)      & \textbf{93.1}              & 87.9 & 89.3  & {91.7}    \\
              & \nicefrac{1}{4} (218)     & {83.0}                & {81.4} & \textbf{87.9}  & {87.4}    \\
              & \nicefrac{1}{8} (109)       & {66.4}               & 64.8   & \textbf{78.8}&  {77.6}    \\
              & \nicefrac{1}{16} (54)     & {51.6}               & 49.6   & \textbf{62.3}  & {60.6}     \\
              & \nicefrac{1}{32} (27) & 44.0 & 30.1 & 47.3 & \textbf{50.3} \\
              \bottomrule
\end{tabularx}
%}
}
%\vspace{-1.5mm}
\caption{$F_1$ scores on the \textsc{dstc8} single-domain data sets. Numbers in brackets denote the training set sizes. The peak scores in each training setup are in bold.}
\label{tab:dstc8Results}
\vspace{-1.5mm}
\end{table}
\begin{table}[!t]
\centering
\def\arraystretch{0.99}
{\small
%\resizebox{0.47\textwidth}{!}
%{
\begin{tabularx}{0.8\textwidth}{r *{2}{Y}}
\toprule
\textbf{Fraction} & \textbf{ConVEx} & \textbf{ConVEx ensemble} \\ \cmidrule(lr){2-3}
1 (8198)        & \textbf{96.0} & 95.8 \\$\nicefrac{1}{2}$ (4099)     & {94.1} & \textbf{94.2}  \\
$\nicefrac{1}{4}$ (2049)     & \textbf{92.5}  & \textbf{92.5}    \\
$\nicefrac{1}{8}$ (1024)     & {90.6}  & \textbf{90.7}   \\
$\nicefrac{1}{16}$ (512)     & {86.4}  & \textbf{88.2}     \\
$\nicefrac{1}{32}$ (256)     & {81.8}  & \textbf{83.9}   \\
$\nicefrac{1}{64}$ (128)     & {76.0}  & \textbf{78.2}     \\
$\nicefrac{1}{128}$ (64)    & {71.7}    & \textbf{73.5}      \\
\bottomrule
\end{tabularx}
}%
%\vspace{-1.5mm}
\caption{Average $F_1$ scores across all slots for \textsc{restaurants-8k} for ConVEx with and without ensembling. The ConVEx ensemble model fine-tunes 3 decoders per slot, and then averages their output scores.}
\label{tab:ensembleResultsFull}
%\vspace{-1.5mm}
\end{table}

%% \ivulic{This 'support set' is not properly defined - can we draw a direct analogy to our terminology? This is what is used at fine-tuning? \matt{yes what we call fine-tuning set in this paper, Hou et al and others call the `support set'.}} 
%\fi

\end{document}